%% file: eccv2020submission.tex
\documentclass[runningheads]{llncs}
\usepackage{graphicx}
\usepackage{comment}
\usepackage{amsmath,amssymb} 
\usepackage{color}
\usepackage[hidelinks]{hyperref}
\input{macro.tex}

\usepackage[width=122mm,left=12mm,paperwidth=146mm,height=193mm,top=12mm,paperheight=217mm]{geometry}

\makeatletter
\def\@fnsymbol#1{\ensuremath{\ifcase#1\or *\or \dagger\or \ddagger\or
   \mathsection\or \mathparagraph\or \|\or **\or \dagger\dagger
   \or \ddagger\ddagger \else\@ctrerr\fi}}
\newcommand{\printfnsymbol}[1]{%
  \textsuperscript{\@fnsymbol{#1}}%
}
\makeatother

\begin{document}
\pagestyle{headings}
\mainmatter
\def\ECCVSubNumber{2295}  

\title{Asynchronous Interaction Aggregation for Action Detection} 

\titlerunning{Asynchronous Interaction Aggregation}
%
\author{Jiajun Tang\inst{1}\thanks{Both authors contributed equally to this work. } \and
Jin Xia\inst{1}\printfnsymbol{1} \and
Xinzhi Mu\inst{1} \and 
Bo Pang\inst{1} \and
Cewu Lu\inst{1}\thanks{The corresponding author is Cewu Lu. Email: lucewu@sjtu.edu.cn}}
\authorrunning{J. Tang and J. Xia and X. Mu and B. Pang and C. Lu}
%
\institute{Shanghai Jiao Tong University, China}
\maketitle

\begin{abstract}
Understanding interaction is an essential part of video action detection. 
We propose the Asynchronous Interaction Aggregation network (AIA) that leverages different interactions to boost action detection. There are two key designs in it: one is the Interaction Aggregation structure (IA) adopting a uniform paradigm to model and integrate multiple types of interaction;
the other is the Asynchronous Memory Update algorithm (AMU) that enables us to achieve better performance by modeling very long-term interaction dynamically without huge computation cost. 
We provide empirical evidence to show that our network can gain notable accuracy from the integrative interactions and is easy to train end-to-end.
Our method reports the new state-of-the-art performance on AVA dataset, with {\it 3.7 mAP} gain ($12.6\%$ relative improvement) on validation split comparing to our strong baseline. The results on dataset UCF101-24 and EPIC-Kitchens further illustrate the effectiveness of our approach. Source code will be made public at: \url{https://github.com/MVIG-SJTU/AlphAction}.

\end{abstract}

\section{Introduction}

\begin{figure}[t]
    \begin{center}
      \includegraphics[width=\linewidth]{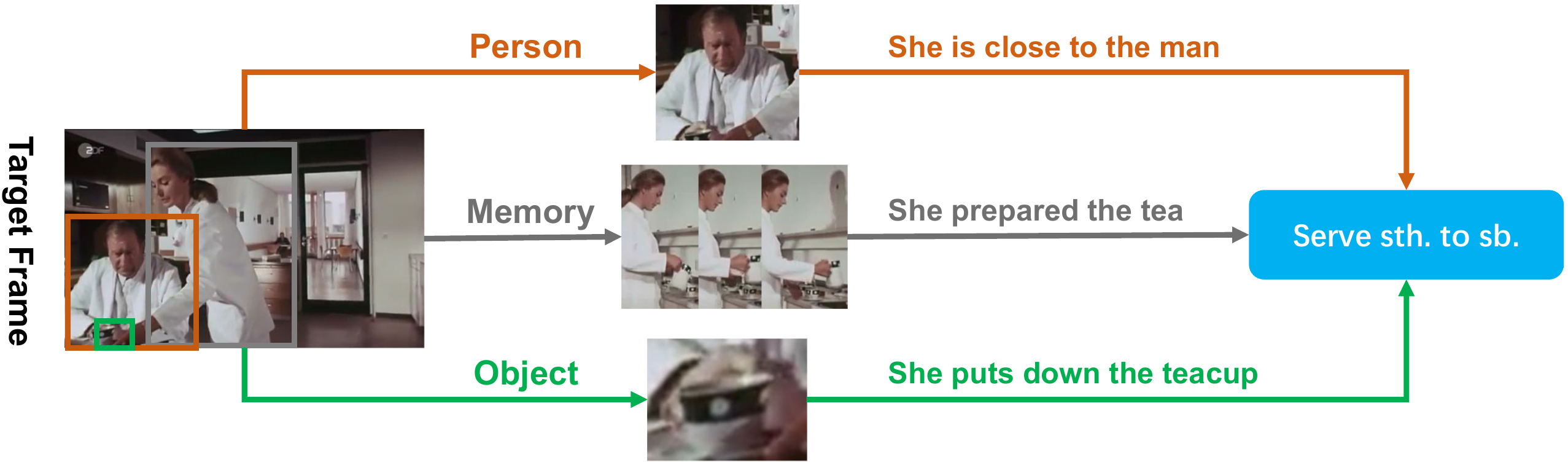}
    \end{center}
        \caption{\textbf{Interaction Aggregation.} In this target frame, we can tell that the women is serving tea to the man with following clues: (1) She is close to the man. (2) She puts down the tea cup before the man. (3) She prepared the tea a few seconds ago. These three clues correspond respectively the person-person, person-object and temporal interactions
        }
    \label{fig:intro}
\end{figure}

\gab{
The task of action detection (spatio-temporal action localization) aims at detecting and recognizing actions in space and time. As an essential task of video understanding, it has a variety of applications such as abnormal behavior detection and autonomous driving. On top of spatial representation and temporal features~\cite{C3D1,P3D,kinetics,SlowFast}, the interaction relationships~\cite{action-transformer,st_graph,temporal_relation,asyn6} are crucial for understanding actions. Take Figure \ref{fig:intro} for example. The appearance of the man, the tea cup as well as the previous movement of the woman help to predict the action of the woman. In this paper, We propose a new framework who emphasizes on the interactions for action detection.

Interactions can be briefly considered as the relationship between the target person and context. Many existing works try to explore interactions in videos, but there are two problems in the current methods: (1) Previous methods such as \cite{action-transformer,hoi} focus on a single type of interaction (eg. person-object). They can only boost one specific kind of actions. Methods such as \cite{structured} intend to merge different interactions, but they model them separately. Information of one interaction can't contribute to another interaction modeling. How to find interactions correctly in video and use them for action detection remains challenging. (2) The long-term temporal interaction is important but hard to track. Methods which use temporal convolution~\cite{C3D1,P3D,SlowFast} have very limited temporal reception due to the resource challenge. Methods such as \cite{lfb} require a duplicated feature extracting pre-process which is not practical in reality. 

In this work, we propose a new framework, the Asynchronous Interaction Aggregation network (AIA), who explores three kinds of interactions (person-person, person-object, and temporal interaction) that cover nearly all kinds of person-context interactions in the video. As a first try, AIA makes them work cooperatively in a hierarchical structure to capture higher level spatial-temporal features and preciser attentions. There are two main designs in our network: the Interaction Aggregation (IA) structure and the Asynchronous Memory Update (AMU) algorithm. 

The former design, IA structure, explores and integrates all three types of interaction in a deep structure. More specifically, it consists of multiple elemental interaction blocks, of each enhances the target features with one type of interaction. These three types of interaction blocks are nested along the depth of IA structure. One block may use the result of previous interactions blocks. Thus, IA structure is able to model interactions precisely using information across different types. 

Jointly training with long memory features is infeasible due to the large size of video data. The AMU algorithm is therefore proposed to estimate intractable features during training. We adopt a memory-like structure to store the spatial features and propose a series of write-read algorithm to update the content in memory: features extracted from target clips at each iteration are written to a memory pool and they can be retrieved in subsequent iterations to model temporal interaction. This effective strategy enables us to train the whole network in an end-to-end manner and the computational complexity doesn't increase linearly with the length of temporal memory features. In comparison to previous solution \cite{lfb} that extracted features in advance, the AMU is much simpler and achieves better performance.

In summary, our key contributions are: (1) A deep IA structure that integrates a diversity of person-context interactions for robust action detection and (2) an AMU algorithm to estimate the memory features dynamically. We perform an extensive ablation study on the AVA \cite{ava} dataset for spatio-temporal action localization task. Our method shows a huge boost on performance, which yields the new state-of-the-art on both validation and test set. We also test our method on dataset UCF101-24 and a segment level action recognition dataset EPIC-Kitchens. Results further validate its generality.

}

\section{Related Works}

\noindent\textbf{Video Classification.}
 Various 3D CNN \cite{C3D1,C3D2,C3D3,C3D4} have been developed to handle video input. To leverage the huge image dataset, I3D \cite{kinetics} has been proposed to benefit from ImageNet\cite{imagenet_cvpr09} pre-training. In \cite{P3D,Spatial-Temporal1,Spatial-Temporal2,Spatial-Temporal3,Spatial-Temporal4}, the 3D kernels in above models are simulated by temporal filters and spatial filters which can significantly decrease the model size. SlowFast Network \cite{SlowFast} introduces a two-stream method \cite{two_stream,two_stream_fusion}. 

\drac{
\noindent\textbf{Spatio-temporal Action Detection.} Action detection is more difficult than action classification because the model needs to not only predict the action labels but also localize the action in time and space. Most of recent approaches \cite{ava,a_better,SlowFast,T-CNN,three_branch} follow the object detection frameworks \cite{FastRCNN,faster} by classifying the features generated by the detected bounding boxes. In contrast to our method, their results depend only on the cropped features. While all the other information is discarded and contributes nothing to the final prediction.   

\noindent\textbf{Attention Mechanism for Videos.} 
The transformer \cite{transformer} consists of several stacked self-attention layers and fully connected layers. Non-Local \cite{nonlocal} concludes that the previous self-attention model can be viewed as a form of classical computer vision method of non-local means \cite{nonlocalmean}. Hence a generic non-local block\cite{nonlocal}  is introduced. This structure enables models to compute the response by relating the features at different time or space, which makes the attention mechanism applicable for video-related tasks like action classification. The non-local block also plays an important role in \cite{lfb} where the model references information from the long-term feature bank via a non-local feature bank operator.
}

\begin{figure}
    \begin{center}
        \includegraphics[width=0.99\linewidth]{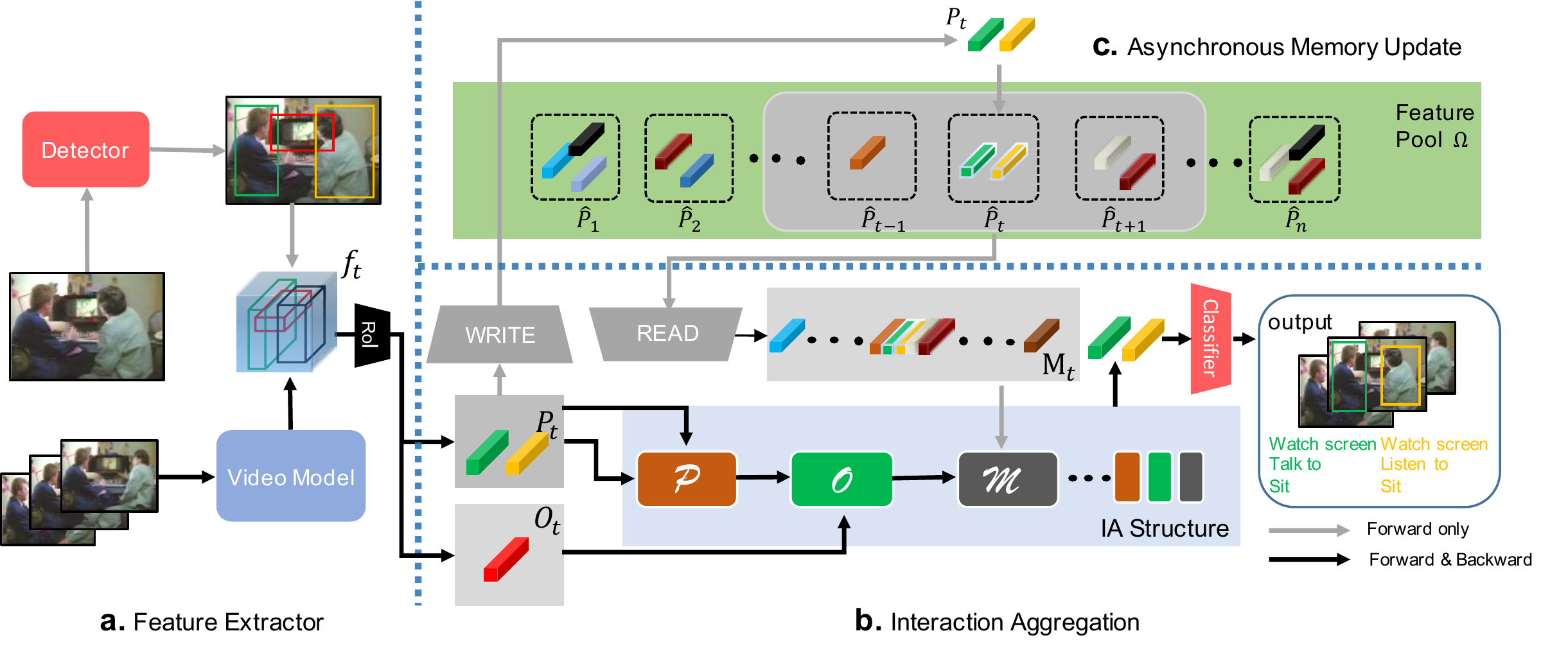}
    \end{center}
       \caption{\textbf{Pipeline of the proposed AIA.}
       \textbf{a}. We crop features of persons and objects from the extracted video features. \textbf{b}. Person features, object features and memory features from the feature pool $\Omega$ in \textbf{c} are fed to IA in order to integrate multiple interactions. The output of IA is passed to the final classifier for predictions. \textbf{c}. Our AMU algorithm reads memory features from feature pool and writes fresh person features to it}
    \label{fig:pipeline}
\end{figure}

\section{Proposed Method}

In this section, we will describe our method that localizes actions in space and time. Our approach aims at modeling and aggregating various interactions to achieve better action detection performance. In Section~\ref{sec:feature}, we describe two important types of instance level features in short clips and the memory features in long videos. In Section~\ref{sec:interaction}, the Interaction Aggregation structure (IA) is explored to gather knowledge of interactions. In Section~\ref{sec:async}, we introduce the Asynchronous Memory Update algorithm (AMU) to alleviate the problem of heavy computation and memory consumption in temporal interaction modeling. The overall pipeline of our method is demonstrated in Figure~\ref{fig:pipeline}.

\subsection{Instance Level and Temporal Memory Features} 
\label{sec:feature}
 
 
 To model interactions in video, we need to find correctly what the queried person is interacted with. Previous works such as \cite{nonlocal} calculate the interactions among all the pixels in feature map. Being computational expensive, these brute-force methods struggle to learn interactions among pixels due to the limited size of video dataset. Thus we go down to consider how to obtain concentrated interacted features. We observe that persons are always interacting with concrete objects and other persons. Therefore, we extract object and person embedding as the instance level feature. In addition, video frames are always highly correlated, thus we keep the long-term person features as the memory features. 

 Instance level features are cropped from the video features. Since computing the whole long video is impossible, we split it to consecutive short video clips $[v_1, v_2, \dots, v_T]$. The $d$-dimensional features of the $t^{th}$ clip $v_t$ is extracted using a video backbone model: $f_t = \mathcal{F}(v_t, \phi_{\mathcal{F}})$ where $\phi_{\mathcal{F}}$ is the parameters.
    
A detector is applied on the middle frame of $v_t$ to get person boxes and object boxes. Based on the detected bounding boxes, we apply RoIAlign to crop the person and object features out from extracted features $f_t$. The person and object features in $v_t$ are denoted respectively as $P_t$ and $O_t$.
    
One clip is only a short session and misses the temporal global semantics. In order to model the temporal interaction, we keep tracks of memory features. The memory features consist of person features in consecutive clips: $M_t = [P_{t-L}, \dots, P_t, \dots, P_{t+L}]$, where $(2L + 1)$ is the size of clip-wise reception field. In practice, a certain number of persons are sampled from each neighbor clip.
    
    
The three features above have semantic meaning and contain concentrated information to recognize actions. With these three features, we are now able to model semantic interactions explicitly.

\subsection{Interaction Modeling and Aggregation}
\label{sec:interaction}
\gab{
    How do we leverage these extracted features? For a target person, there are multiple detected objects and persons. The main challenge is how to correctly pay more attention to the objects or the persons that the target person is interacted with. In this section, we introduce first our Interaction Block that can adaptively model each type of interactions in a uniform structure. Then we describe our Interaction Aggregation (IA) structure that aggregates multiple interactions.   
}
\yelan{
    
    \noindent\textbf{Overview.} Given different human $P_t$, object $O_t$ and memory features $M_t$, the proposed IA structure outputs action features $A_t = \mathcal{E}(P_t, O_t, M_t, \phi_{\mathcal{E}})$, where $\phi_{\mathcal{E}}$ denotes the parameters in the IA structure. $A_t$ is then passed to the final classifier for final predictions. 
    
    The hierarchical IA structure consists of multiple interaction blocks. Each of them is tailored for a single type of interactions. The interaction blocks are deep nested with other blocks to efficiently integrate different interactions for higher level features and more precise attentions.
    
    \noindent\textbf{Interaction Block.} 
    The structure of interaction block is adapted from Transformer Block originally proposed in \cite{transformer} whose specific design basically follows \cite{nonlocal,lfb}. Briefly speaking, one of the two inputs is used as the query and the other is mapped to key and value. Through the dot-product attention, which is the output of the softmax layer in Figure~\ref{fig:blocks} \textbf{a}, the block is able to select value features that are highly activated to the query features and merge them to enhance the query features. There are three types of interaction blocks, P-Block, O-Block and M-Block.
    
    {\it -P-Block:} P-Block models person-person interaction in the same clip. It is helpful for recognizing actions like listening and talking. Since the query input is already the person features or the enhanced person features, we take the key/value input the same as the query input. 
    
    {\it -O-Block:} In O-Block, we aim to distill person-object interactions such as pushing and carrying an object. Our key/value input is the detected object features $O_t$. In the case where detected objects are too many, we sample based on detection scores. Figure~\ref{fig:blocks}\textbf{a} is an illustration of O-Block.

    {\it -M-Block:} Some actions have strong logical connections along the temporal dimension like opening and closing. We model this type of interaction as temporal interactions. To operate this type, we take memory features $M_t$ as key/value input of an M-Block.
    
    \begin{figure}[t]
        \begin{center}
            \includegraphics[width=0.8\linewidth]{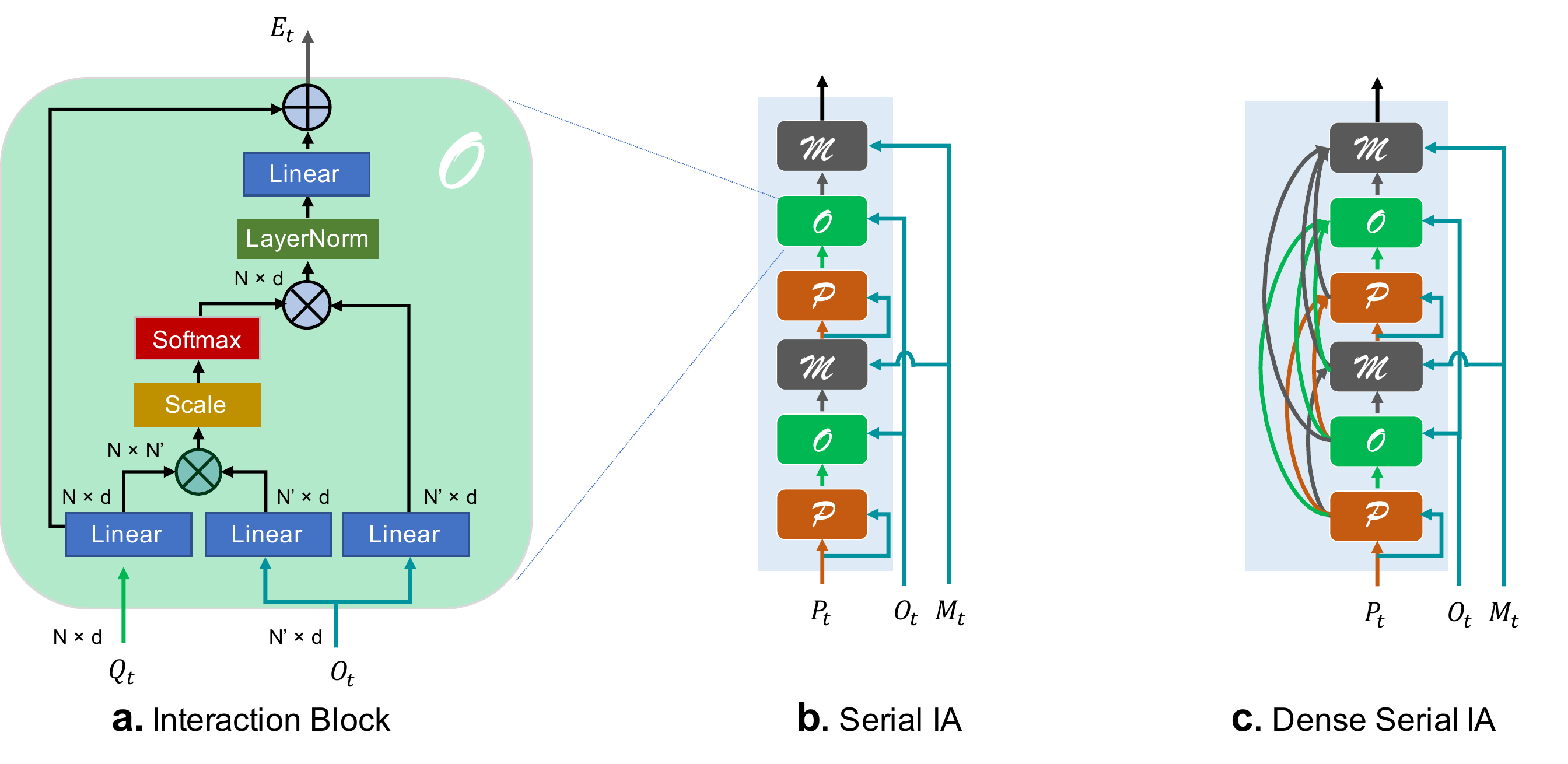}
        \end{center}
           \caption{\textbf{Interaction Block and IA structure.}
           \textbf{a}. The O-Block: the query input is the feature of the target person and the key/value input is the feature of objects. The P-Block and M-Block are similar. \textbf{b}. Serial IA. \textbf{c}. Dense Serial IA
           }
           
        \label{fig:blocks}
    \end{figure}
    
    \noindent\textbf{Interaction Aggregation Structure.} 
    The Interaction Blocks extract three types of interaction. We now propose two IA structures to integrate these different interactions. The proposed IA structures are the naive parallel IA, the serial IA and the dense serial IA. For clarity, we use $\mathcal{P}$, $\mathcal{O}$, and $\mathcal{M}$ to represent the P-Block, O-Block, and M-Block respectively.
    

    {\it -Parallel IA:} A naive approach is to model different interactions separately and merge them at last. As displayed in Figure \ref{fig:vis}\textbf{a}., each branch follows similar structure to \cite{action-transformer} that treats one type of interactions without the knowledge of other interactions. We argue that the parallel structure struggles to find interaction precisely. We illustrate the attention of the last P-Block in Figure \ref{fig:vis}\textbf{c}. by displaying the output of the softmax layer for different persons. As we can see, the target person is apparently watching and listening to the man in red. However, the P block pays similar attention to two men.
    
    \begin{figure}
        \begin{center}
            \includegraphics[width=0.95\linewidth]{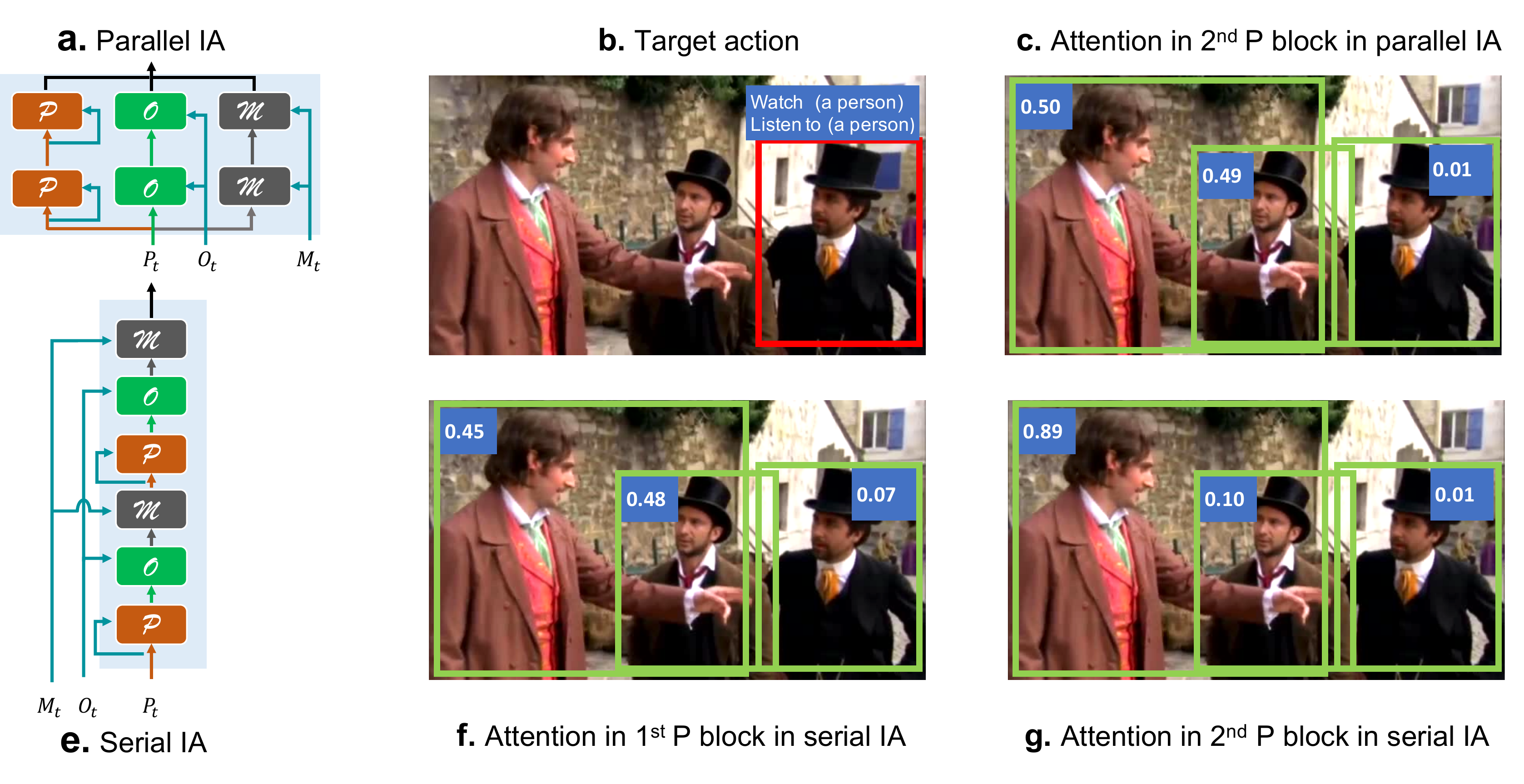}
        \end{center}
           \caption{
           We visualize attention by displaying the output of the softmax layer in P-Block. The original output contains the attention to zero padding person. We remove those meaningless attention and normalize the rest attention to 1
           }
           
        \label{fig:vis}
    \end{figure}


    {\it -Serial IA:} The knowledge across different interactions is helpful for recognizing interactions. We propose the serial IA to aggregate different types of interactions. As shown in Figure \ref{fig:blocks}\textbf{b}., different types of interaction blocks are stacked in sequence. The queried features are enhanced in one interaction block and then passed to an interaction block of a different type. Figure \ref{fig:vis}\textbf{f}. and \ref{fig:vis}\textbf{g}. demonstrate the advantage of serial IA: The first P block can not differ the importance of the man in left and the man in middle. After gaining knowledge from O-block and M-block, the second P-block is able to pay more attention to man in left who is talking to the target person. Comparing to the attention in parallel IA (Figure \ref{fig:vis}\textbf{c}.), our serial IA is better in finding interactions.

    {\it -Dense Serial IA:} In above structures, the connections between interaction blocks are totally manually designed and the input of an interaction block is simply the output of another one. We expect the model to further learn which interaction features to take by itself. With this in mind, we propose the Dense Serial IA extension. In Dense Serial IA, each interaction block takes all the outputs of previous blocks and aggregates them using a learnable weight. Formally, the query of the $i^{th}$ block can be represent as 
    \begin{equation}
        Q_{t,i} = \sum_{j\in\mathbf{C}} W_j \odot E_{t, j},
    \end{equation}
    where $\odot$ denotes the element-wise multiplication, $\mathbf{C}$ is the set of indices of previous blocks, $W_j$ is a learnable $d$-dimenional vector normalized with a Softmax function among $\mathbf{C}$, $E_{t, j}$ is the enhanced output features from the $j^{th}$ block. Dense Serial IA is illustrated in Figure~\ref{fig:blocks}\textbf{c}.
    
    
}

\subsection{Asynchronous Memory Update Algorithm}

\begin{figure}[t]
    \begin{center}
      \includegraphics[width=0.85\linewidth]{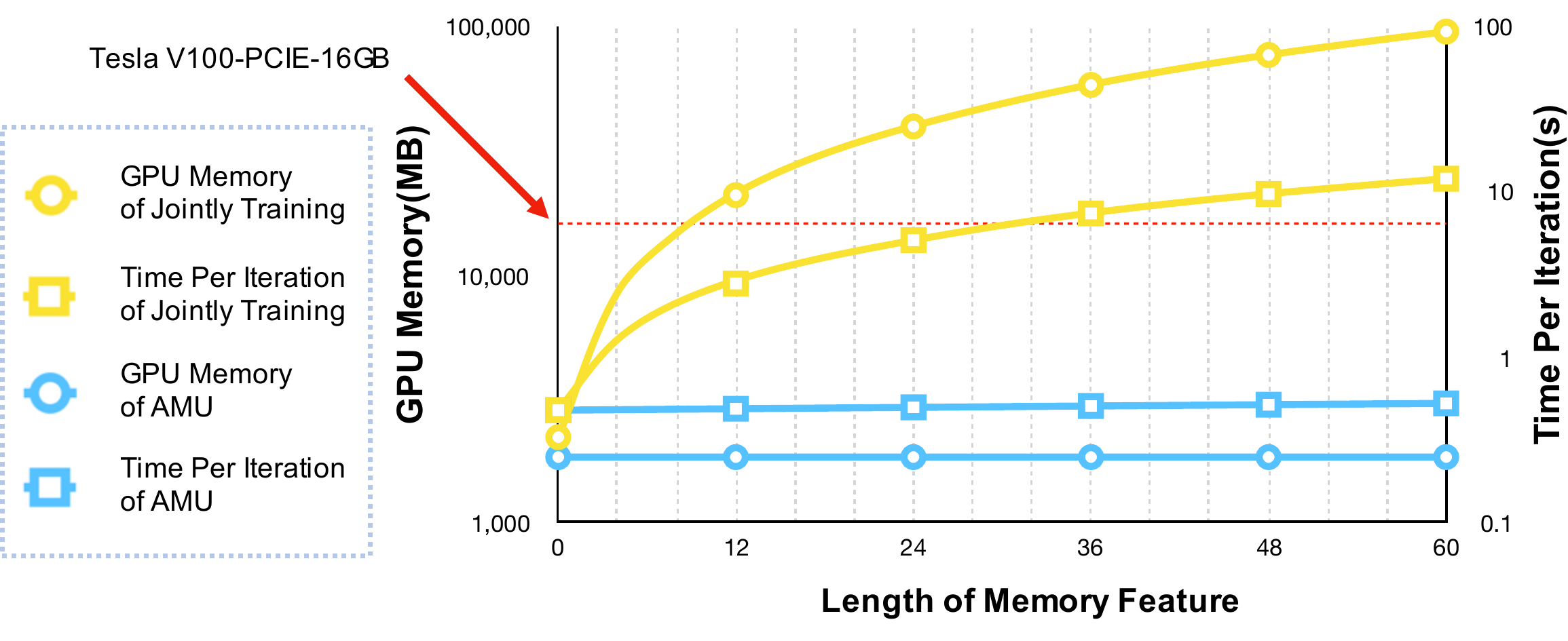}
    \end{center}
       \caption{\textbf{Joint training with memory features is restricted by limited hardware resource.} In this minor experiment, we take a 32-frame video clip with $256\times 340$ resolution as input. The backbone is ResNet-50. During joint training (\textcolor{yellow}{yellow} line), rapidly growing GPU memory and computation time restricted the length of memory features to be very small value (8 in this experiment). With larger input or deeper backbone, this problem will be more serious. Our method (\textcolor{cyan}{cyan} line) doesn't have such problem.
       }
    \label{fig:memory-res}
\end{figure}

Long-term memory features can provide useful temporal semantics to aid recognizing actions. Imagine a scene where a person opens the bottle cap, drinks water, and finally closes the cap, it could be hard to detect opening and closing with subtle movements. But knowing the context of drinking water, things get much easier.

\label{sec:async}

\noindent\textbf{Resource Challenge.} To capture more temporal information, we hope our $M_t$ can gather features from enough clips, however, using more clips will increase the computation and memory consumption dramatically.  Depicted with Figure~\ref{fig:memory-res}, when jointly training, the memory usage and computation consumption increase rapidly as the temporal length of $M_t$ grows. To train on one target person, we must forward and backward $(2L+1)$ video clips at one time, which consumes much more time, and even worse, cannot make full use of enough long-term information due to limited GPU memory. 

\begin{algorithm}
    \fontsize{8}{9.5}\selectfont
    \caption{Training with asynchronous memory update}
    \label{alg:async}
    \begin{algorithmic}[1]
    \renewcommand{\algorithmicrequire}{\textbf{Input:}}
    \renewcommand{\algorithmicensure}{\textbf{Output:}}
    \REQUIRE Video dataset $\mathbf{V}=\{v^{(1)}, v^{(2)}, \dots, v^{(|\mathbf{V}|)}\}$ with $v^{(i)}=[v_1^{(i)}, v_2^{(i)}, \dots, v_{T_i}^{(i)}]$;  \\
    The whole network $\mathcal{N}$, with its parameter $\phi_{\mathcal{N}}$;
    \ENSURE  Optimized network $\mathcal{N}$ with $\phi_{\mathcal{N}}$ for inference.
    \\\textit{// Initialization} :
     \STATE  $\Omega = \{(\hat{P}_t^{(i)}\leftarrow \text{zero vectors}, \delta_t^{(i)}\leftarrow 0)\mid \forall t,i\}$.
     \STATE $err \leftarrow \infty$.
    \\\textit{// Training Process:}
     \FOR {$iter = 1$ to $iter_{max}$}
     \STATE Sample a video clip $v_t^{(i)}$ from dataset $\mathbf{V}$.
     \FOR {$t'=t-L$ to $t+L$}
        \IF{$t'\neq t$}
            \STATE \textbf{\emph{READ}} $\hat{P}_{t'}^{(i)}$ and $\delta_{t'}^{(i)}$ from memory pool $\Omega$.
            \STATE $w_{t'}^{(i)} = \text{min}\{err/\delta_{t'}^{(i)},\delta_{t'}^{(i)}/err\}$.
            \STATE Impose penalty: $\hat{P}_{t'}^{(i)}\leftarrow w_{t'}^{(i)}\hat{P}_{t'}^{(i)}$.
        \ENDIF
     \ENDFOR
     \STATE Extract $P_t^{(i)}$ and $ O_t^{(i)}$ with the backbone in $\mathcal{N}$.
     \STATE Estimated memory features: $\hat{M}_{t}^{(i)}\leftarrow [\hat{P}_{t-L1}^{(i)}, \dots, \hat{P}_{t-1}^{(i)}, P_t^{(i)}, \hat{P}_{t+1}^{(i)}, \dots, \hat{P}_{t+L2}^{(i)}]$.
     \STATE Forward $(P_t^{(i)}, O_t^{(i)}, M_t^{(i)})$ with the head in $\mathcal{N}$ and backward to optimize $\phi_{\mathcal{N}}$.
     \STATE Update $err$ as the output of current loss function.
     \STATE \textbf{\emph{WRITE}} $\hat{P}_t^{(i)}\leftarrow P_t^{(i)}, \delta_t^{(i)}\leftarrow err$ back to $\Omega$.
     \ENDFOR
    \RETURN $\mathcal{N}$, $\phi_{\mathcal{N}}$
    \end{algorithmic}
\end{algorithm}

\noindent\textbf{Insight.} In the previous work \cite{lfb}, they pre-train another duplicated backbone to extract memory features to avoid this problem. 
However, this method make use of frozen memory features, whose representation power can not be enhanced as model training goes. We expect the memory feature can be updated dynamically and enjoy the improvement from parameter update in training process. Therefore, we propose the asynchronous memory update method which can generate effective dynamic long-term memory features and make the training process more lightweight. The details of training process with this algorithm are presented in Algorithm~\ref{alg:async}.  

Inspired by \cite{memnet}, our algorithm is composed of a memory component, the memory pool $\Omega$ and two basic operations, {\it READ} and {\it WRITE}. The memory pool $\Omega$ records memory features. Each feature $\hat{P}_t^{(i)}$ in this pool is an estimated value and tagged with a loss value $\delta_t^{(i)}$. This loss value $\delta_t^{(i)}$ logs the convergence state of the whole network.
Two basic operations are invoked at each iteration of training:

{\it -READ:} At the beginning of each iteration, given a video clip $v_t^{(i)}$ from the $i^{th}$ video, memory features around the target clip are read from the memory pool $\Omega$, which is $[\hat{P}_{t-L}^{(i)}, \dots, \hat{P}_{t-1}^{(i)}]$ and $[\hat{P}_{t+1}^{(i)}, \dots, \hat{P}_{t+L}^{(i)}]$ specifically.

{\it -WRITE:} At the end of each iteration, personal features for the target clip $P_t^{(i)}$ are written back to the memory pool $\Omega$ as estimated memory features $\hat{P}_t^{(i)}$, tagged with current loss value.

{\it -Reweighting:} The features we {\it READ} are written at different training steps. Therefore, some early written features are extracted from the model whose parameters are much different with current ones. Therefore, we impact a penalty factor $w_{t'}^{(i)}$ to discard badly estimated features. We design a simple yet effective way to compute such penalty factor by using loss tag. The difference between the loss tag $\delta_{t'}^{(i)}$ and current loss value is expressed as,
    \begin{equation}
       w_{t'}^{(i)} =  \min \{err/\delta_{t'}^{(i)},\delta_{t'}^{(i)}/err\}, 
    \end{equation}
 which should be very close to 1 when the difference is small. As the network converges, the estimated features in the memory pool are expected to be closer and closer to the precise features and $w_{t'}^{(i)}$ approaches to 1.  

As shown in Figure~\ref{fig:memory-res}, the consumption of our algorithm has no obvious increase in both GPU memory and computation as the length of memory features grows, and thus we can use long enough memory features on current common devices. With dynamic updating, the asynchronous memory features can be better exploited than frozen ones.


\section{Experiments on AVA}
\yelan{
The Atomic Visual Actions(AVA) \cite{ava} dataset is built for spatio-temporal action localization. In this dataset, each person is annotated with a bounding box and multiple action labels at 1 FPS. There are 80 atomic action classes which cover pose actions, person-person interactions and person-object interactions. This dataset contains 235 training movie videos and 64 validation movie videos.

Since our method is originally designed for spatio-temporal action detection, we use AVA dataset as the main benchmark to conduct detailed ablation experiments. The performances are evaluated with official metric frame level mean average precision(mAP) at spatial IoU $\geq 0.5$ and only the top 60 most common action classes are used for evaluation, according to \cite{ava}. 

}

\subsection{Implementation Details}
\noindent\textbf{Instance Detector.}\yelan{
We apply Faster R-CNN \cite{faster} framework to detect persons and objects on the key frames of each clip. A model with ResNeXt-101-FPN \cite{resnext,fpn} backbone from maskrcnn-benchmark \cite{massa2018mrcnn} is adopted for object detection. It is firstly pre-trained on ImageNet \cite{imagenet_cvpr09} and then fine-tuned on MSCOCO \cite{mscoco} dataset. For human detection, we further fine-tune the model on AVA for higher detection precision.

\noindent\textbf{Backbone.}
Our method can be easily applied to any kind of 3D CNN backbone. We select state-of-the-art backbone SlowFast \cite{SlowFast} network with ResNet-50 structure as our baseline model. Basically following the recipe in \cite{SlowFast}, our backbone is pre-trained on Kinetics-700 \cite{kinetics} dataset for action classification task. This pre-trained backbone produces 66.34\% top-1 and 86.66\% top-5 accuracy on the Kinetics-700 validation set.


\noindent\textbf{Training and Inference.}
Initialized from Kinetics pre-trained weights, we then fine-tune the whole model on AVA dataset. The inputs of our network are 32 RGB frames, sampled from a 64-frame raw clip with one frame interval. Clips are scaled such that the shortest side becomes 256, and then fed into the fully convolution backbone. We use only the ground-truth human boxes for training and the randomly jitter them for data augmentation. For the object boxes, we set the detection threshold to 0.5 in order to higher recall. During inference, detected human boxes with a confidence score larger than 0.8 are used. We set $L=30$ for memory features in our experiments.
We train our network using the SGD algorithm with batch size 64 on 16 GPU(4 clips per device). BatchNorm(BN) \cite{batchnorm} statistics are set frozen. We train for 27.5k iterations with base learning rate 0.004 and the learning rate is reduced by a factor 10 at 17.5k and 22.5k iteration. A linear warm-up \cite{linearlr} scheduler is applied for the first 2k iterations. 



}

\subsection{Ablation Experiments}

\noindent\textbf{Three Interactions.}
\yelan{
We first study the importance of three kinds of interactions. For each interaction type, we use at most one block in the experiment. These blocks are then stacked in serial. To evaluate the importance of person-object interaction, we remove the O-Block in the structure. Other interactions are evaluated in the same way.
Table~\ref{tab:three_inter} compares the model performance, where used interaction types are marked with "\checkmark". A backbone baseline without any interaction is also listed in this table.
Overall we observe that removing any of these three type interactions results in a significant performance decrease, which confirms that all these three interactions are important for action detection.
}

\noindent\textbf{Number of Interaction Blocks.}
\yelan{
We then experiment with different settings for the number of interaction blocks in our IA structure. The interaction blocks are nested in serial structure in this experiment. In Table~\ref{tab:num_blocks}, $N\times\{\mathcal{P},\mathcal{M},\mathcal{O}\}$ denotes $N$ blocks are used for each interaction type, with the total number as $3N$. We find that with setting $N=2$ the our method can achieve the best performance, so we use this as our default configuration.
}

\input{abla.tex}

\noindent\textbf{Interaction Order.}
\yelan{
In our serial IA, different type of interactions are alternately integrated in sequential. We investigate effect of different interaction order design in Table~\ref{tab:interact_order}. As shown in this experiment, the performance with different order are quite similar, we thus choose the slightly better one $\mathcal{P}\rightarrow \mathcal{O}\rightarrow \mathcal{M}$ as our default setting.
}

\noindent\textbf{Interaction Aggregation Structure.}
\yelan{
We analyze different IA structure in this part. Parallel IA, serial IA and the dense serial IA extension are compared in Table~\ref{tab:agg_struct}. As we expect, the parallel IA performs much worse than serial structure. With dense connections between blocks, our model is able to learn more knowledge of interactions, which further boosts the performance.
}

\noindent\textbf{Asynchronous Memory Update.}
\yelan{ 
In the previous work LFB \cite{lfb}, the memory features are extracted with another backbone, which is frozen during training. In this experiment we compare our asynchronous memory features with the frozen ones. For fair comparison, we re-implement LFB with SlowFast backbone, and also apply our AMU algorithm to LFB. In Table~\ref{tab:async_memory}, we find that our asynchronous memory features can gain much better performance than the frozen method with nearly half of the parameters and computation cost. We argue that this is because our dynamic features can provide better representation.
}

\noindent\textbf{Comparison to Non-local Attention.}
\yelan{
Finally we compare our interaction aggregation method with prior work non-local block \cite{nonlocal}. Following \cite{SlowFast}, we augment the backbone with a non-local branch, where attention is computed between the person instance features and global pooled features. Since there is no long-term features in the non-local block, we use only $\mathcal{P}$ and $\mathcal{O}$ in this experiment. In Table~\ref{tab:cmp_nl}, we see that our serial IA works significantly better than non-local block. This confirms that our method can better learn to find potential interactions than non-local block.
}

\subsection{Main Results}

\begin{figure}[t]
    \begin{center}
      \includegraphics[width=0.99\linewidth]{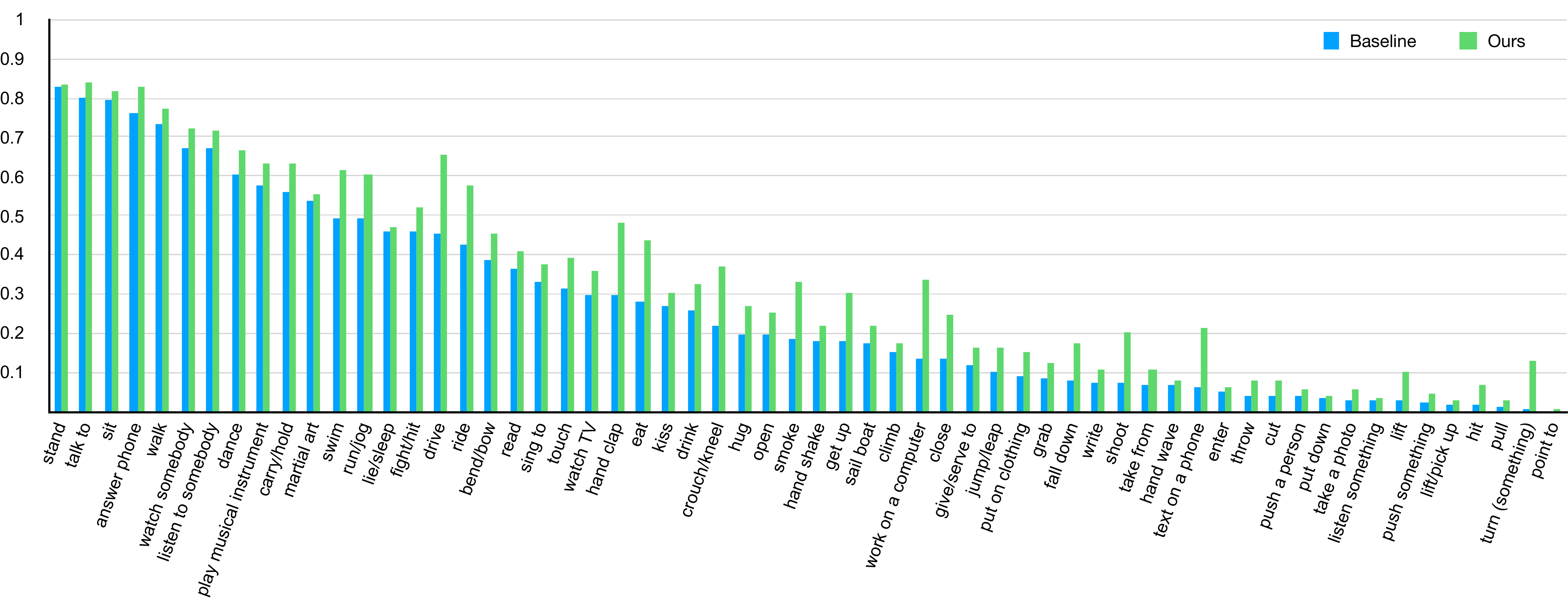}
    \end{center}
      \caption{
      Per category results for the proposed network and the baseline model on the validation set of AVA v2.2
      }
    \label{fig:expriments}
\end{figure}

\yelan{
Finally, we compare our results on AVA with previous methods in Table~\ref{tab:ava_sota}. We show our results on both AVA v2.1 and v2.2. Our method surpass all previous works on both versions. 

The AVA v2.2 dataset, 
is the newer benchmark used in ActivityNet challenge 2019 \cite{caba2015activitynet}. On the validation set, our method reports a new state-of-the-art {\it 33.11 mAP} with one single model, which outperforms the strong baseline SlowFast networks by {\it 3.7 mAP}. On the test split, we train our model on both training and validation splits and use a relative longer scheduler due to more data. With an ensemble of three models with different learning rates and aggregation structures, our method achieves better performance than the winning entry of AVA action detection challenge 2019 (an ensemble with 7 SlowFast networks). The per category results for our method and SlowFast baseline is illustrated in Figure~\ref{fig:expriments}. We can observe the performance gain for each category, especially for those who contain interactions with video context.

We note that  we use a new larger Kinetics-700 for pre-training. 
As shown in Table~\ref{tab:ava_sota}, the backbone we implement in this work has a similar performance to the official model pre-trained with Kinetics-600. This comparison proves that the very most part of performance gain is from our method.
}

\input{sota.tex}

\section{Experiments on UCF101-24}

UCF101-24~\cite{ucf101} is an action detection set with 24 action categories.
In this dataset, the bounding boxes of each action instance are annotated frame by frame. We conduct experiments on the first split of this dataset following previous works. We use the corrected annotations provided by Singh \emph{et al.}~\cite{ROAD}.

\subsection{Implemetnation Details}
Since the bounding boxes are annotated in frame level, we choose to use a ResNet50-C2D proposed in \cite{nonlocal} as our video backbone model. To receive single image as input, we remove all temporal poolings in it. We pre-train it on the Kinetics-400 dataset. Other settings are basically the same as AVA experiments. More implementation details are provided in Supplementary Material. 


\subsection{Quantitative Evaluation}
\input{ucf_tab.tex}
Table~\ref{tab:ucf_cmp} shows the result on UCF101-24 test split in terms of Frame-mAP with 0.5 IOU threshold. As we can see in the table, AIA achieves 3.3\% mAP gain comparing to our C2D baseline. Moreover, with only a lightweight 2D backbone (other works could use some 3D backbone), our method still achieves very competitive results.

\section{Experiments on EPIC-Kitchens}

\begin{table}[t]
\setlength{\tabcolsep}{1mm}
    \centering
    \fontsize{8}{9.5}\selectfont
    \caption{\textbf{EPIC-Kitchens Validation Results}}
    \label{tab:epic_result}
    \begin{tabular}{l@{\hspace{0.3cm}}c@{\hspace{0.3cm}}c@{\hspace{0.3cm}}c@{\hspace{0.3cm}}c@{\hspace{0.3cm}}c@{\hspace{0.3cm}}c}
         & \multicolumn{2}{c}{Verbs} & \multicolumn{2}{c}{Nouns} & \multicolumn{2}{c}{Actions}  \\
          \cline{2-7}
          &top-1&top-5&top-1&top-5&top-1&top-5 \\
        \specialrule{1.5pt}{1pt}{1pt}
          Baradel~\cite{baradel} &40.9 & -&-&-&-&-\\
          LFB NL~\cite{lfb}& 52.4 & 80.8 & 29.3 & 54.9 & 20.8 & 39.8\\
          SlowFast (ours) & 56.8 & 82.8 & 32.3 & 56.7 & 24.1 & 42.0 \\
         \specialrule{0.5pt}{0.5pt}{0.5pt}

            AIA-Parallel & 57.6 & 83.9 & 36.3 & 63.0 & 26.4 & 47.4 \\
          AIA-Serial & 59.2 & 84.2 & \textbf{37.2} & \textbf{63.2} & \textbf{27.7} & \textbf{48.0} \\
          AIA-Dense-Serial & \textbf{60.0} & \textbf{84.6}& \textbf{37.2} & 62.1 & 27.1 & 47.8 \\

    \end{tabular}
\end{table}

\drac{
To demonstrate the generalizability of AIA, we evaluate our method on the segment level dataset EPIC-Kitchens \cite{epic}. 
In EPIC Kitchens, each segment is annotated with one verb and one noun. The action is defined by their combination. 
Following \cite{baradel}, We split the original training set into a new training set and a new validation set. Verb models and noun models are trained separately. Actions are obtained by combining their predictions. 
}

\subsection{Implementation Details}
\drac{
For both verb model and noun model, we use the extracted segment features (global average pooling of $f_t$) as query input for IA structure. Hand features and object features are cropped and then fed into IA to model person-person and person-object interactions. For verb model, the memory features are the segment features. For noun model, the memory features are the object features extracted from object detector feature map, thus the AMU algorithm is only applied to the verb model. Other experimental details different from AVA setting are provided in Supplementary Material.


}

\subsection{Quantitative Evaluation}
We observe from Table~\ref{tab:epic_result} a significant gain for all three tasks. All the variants of AIA outperform the strong baseline SlowFast. Among them, the dense serial IA achieves the best performance for the verbs test, leading to 3.2\% improvement on top-1 score. The serial IA results in 4.9\% for the nouns test and 3.6\% for the action test. 

\section{Conclusion}

In this paper, we present the Asynchronous Interaction Aggregation network and its performance in action detection. Our method reports the new start-of-the-art on AVA dataset. Nevertheless, the performance of action detection and the interaction recognition is far from perfect. The poor performance is probably due to the limited video dataset. Transferring the knowledge of action and interaction from image could be a further improvement for AIA network. 


%
%
\bibliographystyle{splncs04}
\bibliography{egbib}
\end{document}

%% file: macro.tex
\usepackage[dvipsnames]{xcolor}

\definecolor{tangerine}{rgb}{0.95, 0.52, 0.0}
\definecolor{applegreen}{rgb}{0.55, 0.71, 0.0}
\definecolor{brickred}{rgb}{0.8, 0.25, 0.33}

\newcommand{\yelan}[1]{{\leavevmode\color{cyan}Y: #1}}
\newcommand{\gab}[1]{{\leavevmode\color{tangerine}G: #1}}
\newcommand{\drac}[1]{{\leavevmode\color{applegreen}D: #1}}

\renewcommand{\yelan}[1]{#1}
\renewcommand{\gab}[1]{#1}
\renewcommand{\drac}[1]{#1}

\graphicspath{{figs/}}
\usepackage{subcaption}
\usepackage{algorithm}
\usepackage{algorithmic}
\usepackage{booktabs}
\usepackage{multirow}

%% file: abla.tex
\begin{table}[t]
        \centering
        \caption{\textbf{Ablation Experiments.} We use a ResNet-50 SlowFast backbone to perform our ablation study. Models are trained on the AVA(v2.2) training set and evaluated on the validation set. The evaluation metric mAP is shown in \%}
        \fontsize{8}{9.5}\selectfont
        \label{tab:abla}
        \begin{subfigure}[t]{0.25\textwidth}
            \caption{\textbf{3 Interactions}}
            \label{tab:three_inter}
        \end{subfigure}%
        \begin{subfigure}[t]{0.3\textwidth}
            \caption{\textbf{Num of I-Blocks}}
            \label{tab:num_blocks}
        \end{subfigure}%
        \begin{subfigure}[t]{0.45\textwidth}
            \caption{\textbf{Interaction Order}}
        \label{tab:interact_order}
        \end{subfigure}\\
        \begin{subtable}[t]{.25\textwidth}
        \centering
        \begin{tabular}[t]{ccc|c}
              $\mathcal{P}$ & $\mathcal{O}$ & $\mathcal{M}$ & mAP \\
             \specialrule{1.5pt}{0pt}{0pt}
             & & & 26.54 \\
             \hline
             \checkmark & \checkmark & & 28.04 \\
             \checkmark & & \checkmark & 28.86 \\
             & \checkmark & \checkmark & 28.92 \\
             \hline
             \checkmark & \checkmark & \checkmark & \textbf{29.26}
        \end{tabular}
        \end{subtable}%
        \begin{subtable}[t]{0.3\textwidth}
            \centering
            \begin{tabular}[t]{l|c}
                  blocks & mAP \\ 
                 \specialrule{1.5pt}{0pt}{0pt}
                 $1\times\{\mathcal{P}, \mathcal{M}, \mathcal{O}\}$ & 29.26 \\
                 $2\times\{\mathcal{P}, \mathcal{M}, \mathcal{O}\}$  & \textbf{29.64} \\
                 $3\times\{\mathcal{P}, \mathcal{M}, \mathcal{O}\}$  & 29.61 \\
            \end{tabular}
        \end{subtable}%
        \begin{subtable}[t]{0.45\textwidth}
        \centering
        \begin{tabular}[t]{l|c||l|c}
               order & mAP & order & mAP\\ 
             \specialrule{1.5pt}{0pt}{0pt}
   
             $\mathcal{M}\rightarrow \mathcal{O} \rightarrow \mathcal{P}$ & 29.48 & 
             $\mathcal{M}\rightarrow \mathcal{P} \rightarrow \mathcal{O}$  & 29.46 \\
             $\mathcal{O}\rightarrow \mathcal{P} \rightarrow \mathcal{M}$  & 29.51 &
             $\mathcal{O}\rightarrow \mathcal{M} \rightarrow \mathcal{P}$  & 29.53 \\
             $\mathcal{P}\rightarrow \mathcal{M} \rightarrow \mathcal{O}$  & 29.44&
             $\mathcal{P}\rightarrow \mathcal{O} \rightarrow \mathcal{M}$  & \textbf{29.64}\\
        \end{tabular}
        \end{subtable}\\
        \begin{subfigure}[t]{0.25\textwidth}
        \caption{\textbf{IA Structure}}
            \label{tab:agg_struct}
        \end{subfigure}%
        \begin{subfigure}[t]{0.5\textwidth}
        \caption{\textbf{Asynchronous Memory Update}}
            \label{tab:async_memory}
        \end{subfigure}%
        \begin{subfigure}[t]{0.25\textwidth}
        \caption{\textbf{Compare to NL}}
            \label{tab:cmp_nl}
        \end{subfigure}\\
        \begin{subtable}[t]{0.25\textwidth}
        \centering
        \begin{tabular}[t]{l|c}
              structure & mAP \\ 
             \specialrule{1.5pt}{0pt}{0pt}
             Parallel & 28.85 \\
             Serial & 29.64 \\
             Dense Serial & \textbf{29.80} \\
        \end{tabular}
        \end{subtable}%
        \begin{subtable}[t]{0.5\textwidth}
            \centering
            \begin{tabular}[t]{l|ccc}
                  model & params & FLOPs & mAP \\ 
                 \specialrule{1.5pt}{0pt}{0pt}
                 Baseline & $1.00\times$ & $1.00\times$ & 26.54\\
                 \hline
                 LFB(w/o AMU) & $2.18\times$ & $2.12\times$ & 27.02 \\ 
                 LFB(w/ AMU) & $1.18\times$ & $1.12\times$ & 28.57  \\
                 \hline
                 IA(w/o AMU) & $2.35\times$ & $2.15\times$ & 28.07 \\ 
                 IA(w/ AMU) & $1.35\times$ & $1.15\times$ & 29.64 \\
            \end{tabular}
        \end{subtable}%
        \begin{subtable}[t]{0.25\textwidth}
            \centering
            \begin{tabular}[t]{l|c}
                  model & mAP \\ 
                 \specialrule{1.5pt}{0pt}{0pt}
                 Baseline & 26.54 \\ \hline
                 +NL & 26.85 \\
                 +IA(w/o $\mathcal{M}$) & \textbf{28.23} \\ 
            \end{tabular}
        \end{subtable}
    \end{table}

%% file: sota.tex
\begin{table}[t]
        \centering
        \caption{\textbf{Main results on AVA dataset.}  In this table, we display our best result with both backbone ResNet-50(R-50) and ResNet-101(R-101). ``*" indicates that the result is tested in multi-scale. 
        The input sizes are shown as the frame number and corresponding sample rate.
        Two R-101 backbone baseline re-implemented in this work are also displayed as ``ours" for comparison. 
        } 
        \label{tab:ava_sota}
        \fontsize{8}{9.5}\selectfont
        \begin{subfigure}[t]{0.45\textwidth}
        \caption{\textbf{Comparison on AVA v2.1.}}
        \label{tab:ava_2.1}
        \end{subfigure}%
        \begin{subfigure}[t]{0.55\textwidth}
        \caption{\textbf{Comparison on AVA v2.2.}}
        \label{tab:ava_2.2}
        \end{subfigure}\\
        \begin{subtable}[t]{0.45\textwidth}
        \centering
        \begin{tabular}[t]{l|c|c|c}
              model & input & pretrain & val \\ 
             \specialrule{1.5pt}{0pt}{0pt}
             SlowFast \cite{SlowFast} & $32\times 2$ & K400 & 26.3 \\
             LFB \cite{lfb} & $32\times 2$ & K400 & 27.7 \\
             I3D \cite{a_better} & $64\times 1$ & K600 & 21.9\\
             SlowFast \cite{SlowFast} & $32\times 2$ & K600 & 28.2 \\
             SlowFast (ours) & $32\times 2$ & K700 & 28.1 \\ 
             \hline
             AIA R50 & $32\times 2$ & K700 & 28.9 \\ 
             AIA R101 & $32\times 2$ & K700 & \textbf{31.2} \\
        \end{tabular}
        \end{subtable}%
        \begin{subtable}[t]{0.55\textwidth}
        \centering
        \begin{tabular}[t]{l|c|c|cc}
              model & input & pretrain & val & test \\ 
             \specialrule{1.5pt}{0pt}{0pt}
             SlowFast \cite{SlowFast} & $32\times 2$ & K600 & 29.1 & -\\
             SlowFast \cite{SlowFast} & $64\times 2$ & K600 & 29.4 & -\\
             SlowFast*, 7 ens. \cite{SlowFast} & - & K600 & - & 34.25\\
             SlowFast (ours) & $32\times 2$ & K700 & 29.3 & -\\
             \hline
             AIA R50 & $32\times 2$ & K700 & 29.80 & -\\
             AIA R101 & $32\times 2$ & K700 & 32.26 & -\\
             AIA R101* & $32\times 2$ & K700 & \textbf{33.11} & \textbf{32.25}\\
             AIA R101*, 3 ens. & - & K700 & - & \textbf{34.42}\\
            \end{tabular}
        \end{subtable}
    \end{table}

%% file: ucf_tab.tex
\begin{table}[t]
    \centering
    \fontsize{8}{9.5}\selectfont
    \caption{\textbf{Results on UCF101-24 Split1}}
    \label{tab:ucf_cmp}
    \begin{tabular}{l@{\hspace{0.4cm}}c@{\hspace{0.4cm}}l@{\hspace{0.4cm}}c@{\hspace{0.4cm}}l@{\hspace{0.4cm}}c}
        \specialrule{0.5pt}{1.5pt}{1.5pt}
        method & mAP & method & mAP & method & mAP\\
        \specialrule{1.5pt}{1.5pt}{1.5pt}
        T-CNN~\cite{T-CNN} & 41.4 &STEP~\cite{STEP} & 75.0 & C2D (ours) & 75.5\\
        Peng \emph{et al.}~\cite{ucfmulti} & 65.7 &Gu \emph{et al.}~\cite{ava} & 76.3 & AIA-Serial & 78.5\\
        ACT~\cite{ACT} & 69.5 &Zhang \emph{et al.}~\cite{structured} & 77.9 & AIA-Dense-Serial & \textbf{78.8} \\
        \specialrule{0.5pt}{1.5pt}{1.5pt}
    \end{tabular}
\end{table}